\pdfoutput=1

\documentclass[11pt]{article}

\usepackage[final]{acl}

\usepackage{times}
\usepackage{latexsym}

\usepackage[T1]{fontenc}

\usepackage[utf8]{inputenc}

\usepackage{microtype}

\usepackage{inconsolata}

\usepackage{graphicx}

\usepackage{algorithm}
\usepackage{algorithmic}
\usepackage{multicol}
\usepackage{multirow}
\usepackage{booktabs}
\usepackage{amsfonts}
\usepackage{colortbl}
\newcommand{\secref}[1]{\S\ref{#1}}

\definecolor{darkred}{RGB}{200, 0, 0}

\title{BlockPruner: Fine-grained Pruning for Large Language Models}

\author{
Longguang Zhong\textsuperscript{1}, Fanqi Wan\textsuperscript{1}, Ruijun Chen\textsuperscript{1}, Xiaojun Quan\thanks{Corresponding author}\textsuperscript{1}, Liangzhi Li\textsuperscript{2}\\
\textsuperscript{1}School of Computer Science and Engineering, Sun Yat-sen University \\
\textsuperscript{2}Meetyou AI Lab \\
\texttt{\{zhonglg5,wanfq,chenrj8\}@mail2.sysu.edu.cn}\\
\texttt{quanxj3@mail.sysu.edu.cn},
\texttt{liliangzhi@xiaoyouzi.com} \\ 
}

\begin{document}
\maketitle
\begin{abstract}
With the rapid growth in the size and complexity of large language models (LLMs), the costs associated with their training and inference have escalated significantly. Research indicates that certain layers in LLMs harbor substantial redundancy, and pruning these layers has minimal impact on the overall performance. While various layer pruning methods have been developed based on this insight, they generally overlook the finer-grained redundancies within the layers themselves.
In this paper, we delve deeper into the architecture of LLMs and demonstrate that finer-grained pruning can be achieved by targeting redundancies in multi-head attention (MHA) and multi-layer perceptron (MLP) blocks. We propose a novel, training-free structured pruning approach called BlockPruner. Unlike existing layer pruning methods, BlockPruner segments each Transformer layer into MHA and MLP blocks. It then assesses the importance of these blocks using perplexity measures and applies a heuristic search for iterative pruning.
We applied BlockPruner to LLMs of various sizes and architectures and validated its performance across a wide range of downstream tasks. Experimental results show that BlockPruner achieves more granular and effective pruning compared to state-of-the-art baselines.
\end{abstract}

\section{Introduction}
Large language models (LLMs) \citep{zhao2023survey, minaee2024large} have demonstrated outstanding performance across a diverse array of natural language processing tasks.~However, their growing size and complexity have led to substantial computational demands and increased memory usage, creating obstacles for deployment in resource-constrained environments.
Model compression techniques \citep{gao2020discrete, li2023losparse, wang2024model}  have emerged as a promising solution to address the challenges of deploying large, computationally intensive models. 
These techniques aim to transform large models into more compact versions that require less storage and execute with lower latency, while minimizing performance degradation.
Model compression methods typically involve knowledge distillation \citep{huang2022context, gu2024minillm}, quantization \citep{NEURIPS2022_adf7fa39, dettmers2023qlora}, and pruning \citep{ouderaa2024the, ashkboos2024slicegpt}. In this study, we primarily focus on pruning, a technique that can be combined with these other methods to achieve more effective and efficient compression.

\begin{figure}
    \centering
    \includegraphics[width=0.9\linewidth]{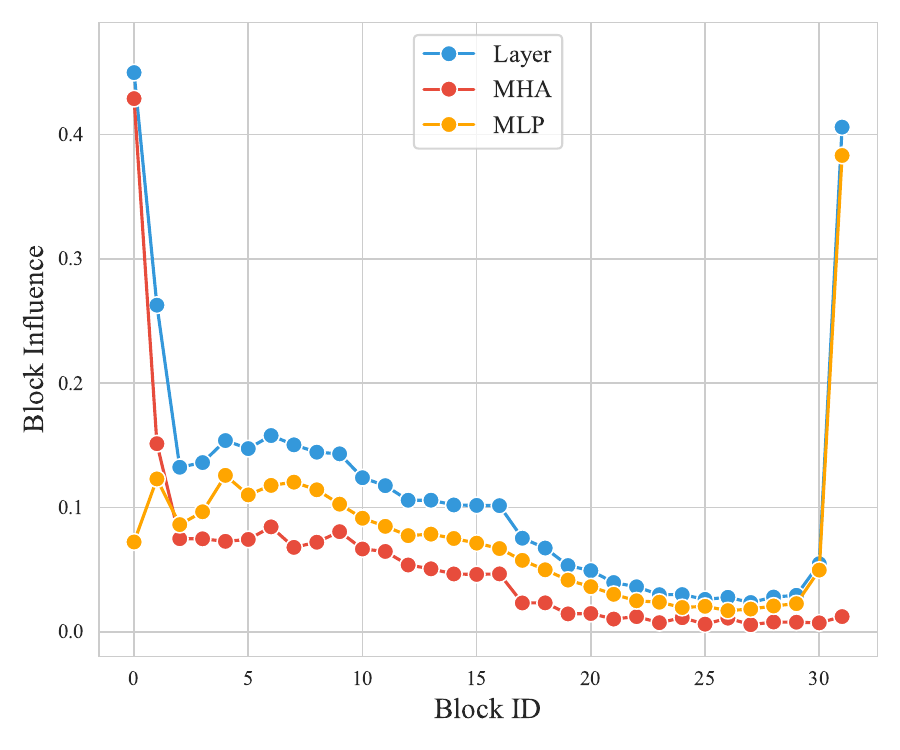}
               \vspace{-0.15cm}
    \caption{Block Influence (BI) scores \citep{men2024shortgpt} for the Llama2-7B model \citep{touvron2023llama2} computed at both layer and block levels, where blocks/layers with lower BI scores indicate less importance. The model has 32 Transformer layers, each containing one MHA and one MLP block, totaling 64 blocks. Block-level BI scores are generally lower than layer-level scores, indicating finer-grained redundancies.
    }
            \vspace{-0.45cm}
    \label{fig:diff_grain_bi}
\end{figure}

Recent research on layer redundancy has shown that LLMs contain a substantial number of redundant layers \citep{yang2024laco, men2024shortgpt, chen2024compressing}. Removing these layers does not severely impact the model's performance. To quantify this redundancy, researchers have investigated various similarity-based measurement methods and developed corresponding pruning strategies, including layer merging \citep{yang2024laco} and layer removal \citep{men2024shortgpt}. 
These methods not only maintain the original width of the model architecture and avoid introducing additional structures, but also demonstrate superior performance.
Furthermore, \citet{gromov2024unreasonable} posited that this observed redundancy may be intrinsically linked to the residual structure \citep{he2016deep} inherent in the Transformer architecture.
Building on this intuition and recognizing that Transformer layers can be further subdivided into smaller residual blocks, namely multi-head attention (MHA) and multi-layer perceptron (MLP)\footnote{In this work, unless otherwise specified, we refer to a block as one of the two sublayers: MHA or MLP.}, we hypothesize that fine-grained block redundancies could exist within LLMs.
Consequently, we conducted a preliminary experiment to assess the significance of blocks at varying granularities. Specifically, we sampled 32 instances from the Alpaca dataset \citep{alpaca} and employed the Block Influence (BI) metric \citep{men2024shortgpt} to evaluate blocks at layer and block levels, as depicted in Figure \ref{fig:diff_grain_bi}. The results reveal that block-level BI scores are generally lower than layer-level BI scores, indicating that fine-grained redundancies at the block level are more significant within the model.

Building on these findings, we argue that finer-grained pruning can be effectively implemented in LLMs. Therefore, we introduce BlockPruner, a novel, training-free structured pruning approach. Unlike existing methods that focus on entire layers, BlockPruner segments each Transformer layer into MHA and MLP blocks. It then evaluates the importance of these blocks using perplexity measures and applies a heuristic search for iterative pruning.

To validate the effectiveness of our method, we applied BlockPruner to six LLMs of varying sizes and architectures, and evaluated their performance using five representative benchmarks. Our experimental results demonstrate that BlockPruner provides more granular and effective pruning compared to state-of-the-art baselines. Additionally, we performed a series of analytical experiments to investigate the impact of block type, block importance metrics, and data on pruning effectiveness. Our findings confirm that LLMs contain substantial redundancies at the block level compared to the layer level, demonstrating that fine-grained pruning is more effective and appropriate than layer-based approaches for compressing these models.

\section{Related Work}
Pruning is a well-established technique to compress and accelerate neural networks by removing superfluous weights or structures within models. 
Pruning methods can be broadly categorized into unstructured pruning and structured pruning.

\paragraph{Unstructured pruning.} Unstructured pruning targets individual weights, eliminating redundant connections in neural networks by setting the corresponding weights to zero. For instance, 
SparseGPT \citep{frantar2023sparsegpt} formulates pruning as a layer-wise sparse regression problem, approximately solving it via a sequence of efficient Hessian updates and weight reconstructions. 
Wanda \citep{sun2024a} computes the importance score of each weight based on the product of the magnitude of each weight and the norm of the corresponding input activation, identifying and removing weights with lower importance scores.
OWL \citep{yin2024outlier} identifies the correlation between pruning efficacy and the retention ratio of outliers, assigning different sparsity ratios to each layer based on the observed outlier ratio.
RIA \citep{zhang2024plugandplay} introduces a metric that considers both weight and activation information, utilizing a permutation strategy for the input channels of weight matrices to enhance pruning performance. 
BESA \citep{xu2024besa} adopts a layer-wise pruning strategy, independently pruning each Transformer layer to minimize the reconstruction error between the outputs of pruned and dense Transformer layers, which avoids accumulating errors across layers.

\paragraph{Structured pruning.} Structured pruning focuses on broader network structures, such as neurons, attention heads, or even entire modules. LLM-Pruner \citep{ma2023llmpruner} utilizes gradient information to identify interdependent structures within LLMs, pruning the least important groups and subsequently using Low-Rank Adaptation (LoRA) \citep{hu2022lora} to restore the performance of pruned models. 
LoRAPrune \citep{zhang2023pruning} estimates the importance of pre-trained weights using LoRA gradients, iteratively removing redundant channels in the weight matrices and recovering the pruned models' performance through fine-tuning.
Sheared-LLaMA \citep{xia2024sheared} learns a set of pruning masks to extract a sub-network with the specified target structure from the source model, employing a dynamic batch loading algorithm to adjust the data proportion of each domain based on the loss reduction rate in different domains. SliceGPT \citep{ashkboos2024slicegpt} introduces the concept of computational invariance, achieving compression by removing rows or columns corresponding to smaller principal components in the weight matrix. LaCo \citep{yang2024laco} proposes a concise layer pruning approach, reducing model size by merging layers while maintaining the overall model structure. ShortGPT \citep{men2024shortgpt} introduces a metric for measuring layer importance, achieving model compression by removing redundant layers. 

Although unstructured pruning can maintain performance at higher pruning ratios, it often requires additional hardware or library support, making model acceleration impractical.
Current structured pruning methods typically require retraining the model after pruning to avoid performance collapse. While layer pruning techniques like LaCo eliminate the need for additional retraining, their disregard for fine-grained block redundancy makes it challenging to avoid significant performance loss.

Concurrent and independent of our research, FINERCUT \citep{zhang2024finercut} also presents a fine-grained block pruning algorithm. 
However, their study does not delve into the rationale behind treating Transformer layers as two distinct sublayers for pruning purposes. 
In contrast, we began by conducting preliminary experiments that unveiled the fine-grained block redundancy within Transformer models. This discovery led us to propose the concept of minimal residual blocks. 
Additionally, we explored how pruning different types of blocks impacts model performance.
While FINERCUT assesses block importance by comparing the similarity between the output logits of the original and pruned models, this metric may fall short in ensuring that the pruned model produces coherent and semantically meaningful text, as it disregards semantic nuances. 
In our approach, we evaluate block importance using the perplexity of the pruned model, a metric that more effectively captures the fluency and quality of its outputs. To further support our perspective, we present a detailed comparison of these two metrics in Appendix \ref{sec:ppl_vs_js}.

\begin{figure}
    \centering
    \includegraphics[width=0.99\linewidth]{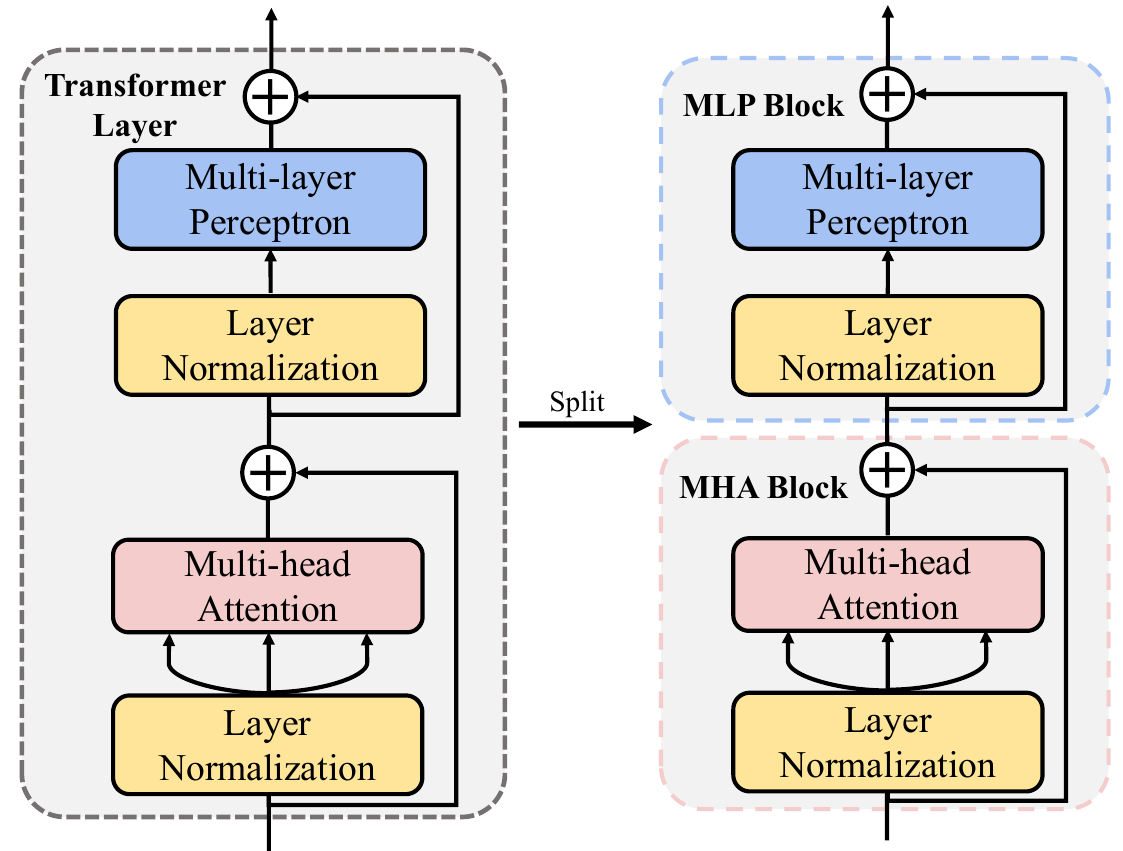}
    \caption{Illustration depicting that a Transformer layer can be subdivided into two residual blocks.}
    \label{fig:residual_block}
          \vspace{-0.3cm}
\end{figure}

\section{Methodology}

\begin{figure*}[!t]
    \centering
    \includegraphics[width=0.97\linewidth]{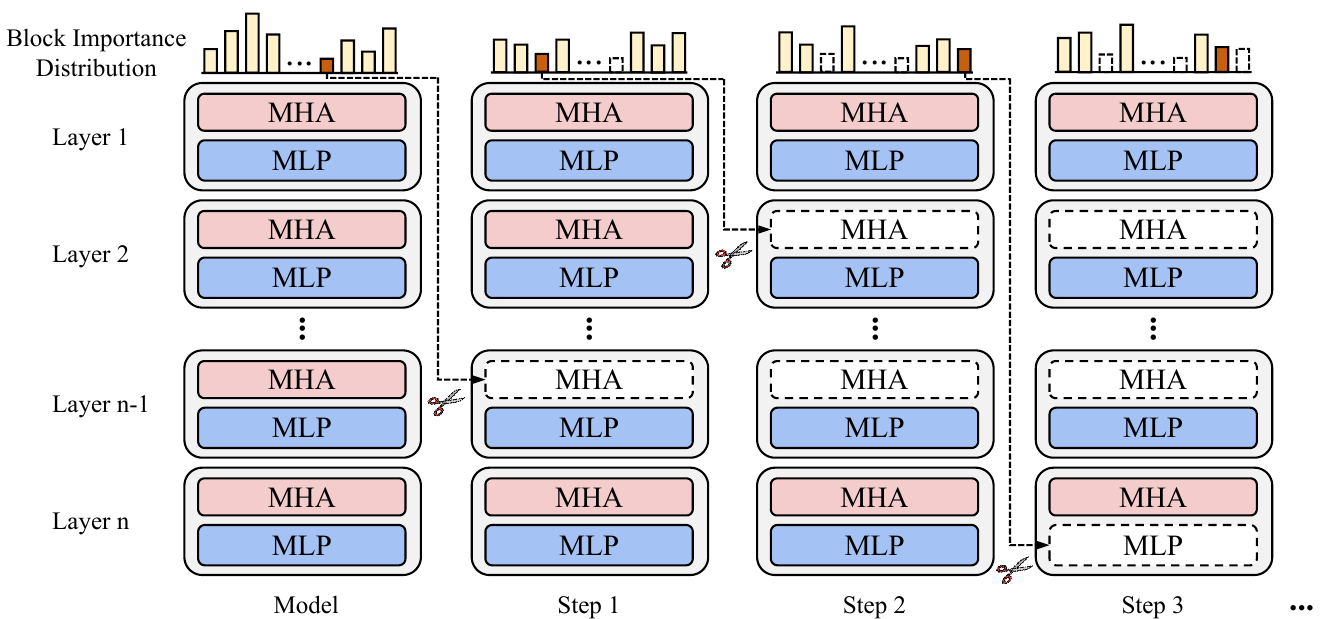}
               \vspace{-0.1cm}
    \caption{Overview of our BlockPruner. We iteratively calculate the importance score for each block (MHA or MLP) to obtain the block importance distribution, and subsequently remove the block with the lowest importance.}
    \label{fig:overview}
               \vspace{-0.3cm}
\end{figure*}

The proposed fine-grained block pruning method (BlockPruner) is depicted in Figure \ref{fig:overview}. It begins by decomposing each Transformer layer into two minimal residual blocks (\secref{sec:mini_resi_block}). We then evaluate the importance of each block by leveraging perplexity for our iterative block pruning framework (\secref{sec:block_impo_metric}). Finally, we iteratively prune the block with the lowest importance (\secref{sec:fgbp}).

\subsection{Minimal Residual Block}
\label{sec:mini_resi_block}
Most contemporary LLMs \citep{brown2020language, touvron2023llama, touvron2023llama2} are built upon the GPT architecture \citep{radford2019language}, which constitutes a decoder-only model comprising multiple Transformer layers, an embedding layer, and a language model head. As depicted in Figure \ref{fig:residual_block}, each Transformer layer can be decomposed into two primary residual blocks: the multi-head attention (MHA) block and the multi-layer perceptron (MLP) block.

Formally, consider the input hidden states of the $i$th Transformer layer, denoted as ${X}_{i-1} \in \mathbb{R}^{n \times d}$, where $n$ represents the length of the input sequence, and $d$ represents the hidden layer dimension of the model. The computational process within the $i$th Transformer layer can be represented as follows:
\begin{equation}\label{eqn:mha}
X_i'=\mathrm{MHA}(\mathrm{LN}(X_{i-1}))+X_{i-1},
\end{equation}
\begin{equation}\label{eqn:mlp}
X_i=\mathrm{MLP}(\mathrm{LN}(X_i'))+X_i'.
\end{equation}
Here, $\mathrm{LN}(\cdot)$ denotes the layer normalization module and $X_i' \in \mathbb{R}^{n \times d}$ represents the intermediate hidden states after the MHA block.

Equations (\ref{eqn:mha}) and (\ref{eqn:mlp}) indicate that both types of residual blocks can be abstracted into the same computational formula. Hence, we argue that treating MLP and MHA blocks as the minimal units for pruning is a reasonable choice, which is substantiated by our subsequent experimental results.

\subsection{Block Importance}
\label{sec:block_impo_metric}
While previous layer pruning methods \citep{men2024shortgpt, chen2024compressing} rely solely on the similarity between layer inputs and outputs to measure layer importance, we argue that this approach overlooks the layer's contribution to the overall model performance, while our metric considers its broader impact on the final output.
To address the drawback, we introduce \emph{perplexity} as a measure of block importance. 
Specifically, we determine the importance score of each block by masking it and then computing the perplexity of the new model on a given dataset.  Intuitively, a block with the lowest importance score indicates that its removal results in minimal performance degradation.
This method more effectively captures each block's overall impact on the model's performance, thereby more accurately reflecting its significance.

Mathematically, perplexity is defined as the exponential of the average negative log-likelihood of a sequence of words. Given a sequence of words $w_1,\ldots,w_n$ and a language model that predicts the probability $p_\theta(w_i|w_{<i})$ for each word $w_i$, the perplexity $\mathrm{PPL}$ is calculated as:
\begin{equation}\label{eqn:ppl}
\setlength{\abovedisplayskip}{8pt}
\setlength{\belowdisplayskip}{8pt} 
\mathrm{PPL}=\mathrm{exp}(-\frac{1}{n}\sum_{i=1}^n \mathrm{\log} p_\theta(w_i|w_{<i})),
\end{equation}
where $p_\theta(w_i|w_{<i})$ denotes the probability of word $w_i$ given the preceding words in the sequence.

\subsection{Iterative Search for Block Pruning}
\label{sec:fgbp}
Unlike existing layer pruning techniques, which indiscriminately remove entire Transformer layers, we propose a novel fine-grained pruning strategy. This strategy selectively prunes MHA or MLP blocks based on their defined importance. 
By employing this finer-grained pruning approach, we aim to better preserve the critical components and capabilities of the model while aggressively removing the less significant blocks.

For an LLM $\mathcal{M}$ with $L$ layers, we first divide them into $2L$ blocks, consisting of MHA and MLP blocks. Then, we perform iterative pruning search on a calibration dataset $\mathcal{C}$ to sequentially prune $K$ blocks. The steps are outlined as follows:

\textbf{Step 1: Mask Block}. For each block $B_i$ (MHA or MLP) in $\mathcal{M}$, we generate a modified model $\hat{\mathcal{M}}$ by masking out this block. 

\textbf{Step 2: Calculate Importance}. We compute the perplexity $P_i$ for the modified model $\hat{\mathcal{M}}$ on the calibration dataset $\mathcal{C}$ as the importance score for the masked block $B_i$.

\textbf{Step 3: Sort and Prune}. After computing the importance scores for all blocks, we sort these scores and remove the block with the lowest importance score from $\mathcal{M}$ to create a new model.

\textbf{Step 4: Iterate}. The aforementioned steps are iteratively repeated until $K$ blocks are removed.

By iteratively removing the blocks with the lowest importance scores, we aim to prune the LLM while minimizing performance degradation on the calibration dataset $\mathcal{C}$. 
This fine-grained block pruning approach provides a more targeted method for pruning LLMs compared to traditional layer-level pruning techniques, thereby facilitating more efficient model compression while better preserving the model's performance. The detailed procedure for this pruning process is outlined in Algorithm \ref{algo1}.

\begin{algorithm}[t]
\caption{Iterative Block Pruning}
\label{algo1}
\begin{algorithmic}[1]
\renewcommand{\algorithmicrequire}{\textbf{Input:}}
\renewcommand{\algorithmicensure}{\textbf{Output:}}
\REQUIRE Model \(\mathcal{M}\) with $L$ layers, calibration dataset $\mathcal{C}$, number of blocks to remove $K$
\ENSURE Pruned model \(\mathcal{M}^*\)
\STATE \(\mathcal{M}_0\) $\gets$ \(\mathcal{M}\)
\STATE Split the model \(\mathcal{M}_0\) into $2L$ blocks
\FOR{$j = 1$ to $K$}
\FOR{$i=1$ to $2L-j+1$}
    \STATE Create model $\hat{\mathcal{M}}$ by masking block $B_i$;
    \STATE Compute the perplexity $P_i$ for $\hat{\mathcal{M}}$ on the calibration dataset $\mathcal{C}$;
\ENDFOR
\STATE Sort the blocks based on their perplexities;
    \STATE Remove the block with the lowest perplexity from \(\mathcal{M}_{j-1}\) and obtain \(\mathcal{M}_j\);
\ENDFOR
\STATE \(\mathcal{M}^*\) $\gets$ \(\mathcal{M}_K\)
\RETURN Pruned model \(\mathcal{M}^*\)
\end{algorithmic}
\end{algorithm}

\vspace{-0.8cm}
\section{Experiments}
In this section, we first introduce the experimental setups and then present the main results.
\subsection{Experimental Setups}
\paragraph{Models.}
To validate the widespread effectiveness of our pruning method, we experiment with three series of models: Llama2 \citep{touvron2023llama2}, Baichuan2 \citep{yang2023baichuan}, and Qwen1.5 \citep{qwen}. These models share analogous architectures as described in equations (\ref{eqn:mha}) and (\ref{eqn:mlp}). Due to computational constraints, we employ 7B and 13B models for Llama2 and Baichuan2, respectively, and 7B and 14B models for Qwen1.5.
\paragraph{Baselines.}
\begin{table*}[!t]
    \centering
\resizebox{0.99\textwidth}{!}{
    \begin{tabular}{llcccccccc}
    \toprule
        \textbf{Model} & \textbf{Method} & \textbf{Ratio (\%)} & \textbf{PPL ($\downarrow$)} &\textbf{PIQA} & \textbf{WinoGrande} & \textbf{HellaSwag} & \textbf{ARC-e} & \textbf{ARC-c} & \textbf{Avg. Score} \\ 
    \midrule
        \multirow{6}{*}{\textbf{Llama2-7B}} & Dense & 0 & 5.47 & 79.05  & 69.06  & 75.99  & 74.54  & 46.16  & 68.96   \\ 
          ~ & SliceGPT & 21.45 & 30.74 & 72.42  & 59.91  & 56.04  & \textbf{63.64}  & 37.12  & 57.83   \\ 
          ~ & LaCo & 21.02 & 50.39 & 68.34  & 60.46  & 54.08  & 55.39  & 35.84  & 54.82   \\ 
          ~ & RM & 21.02 & 676.80 & 54.46  & 49.25  & 29.22  & 34.43  & 22.53  & 37.98   \\ 
          ~ & ShortGPT & 21.02 & 18.45 & 70.24  & \textbf{65.90}  & 62.63  & 56.06  & 36.09  & 58.18   \\ 
          \rowcolor[gray]{.93} \cellcolor{white} ~ & \cellcolor{white}BlockPruner & \cellcolor{white}21.99 &  \textbf{11.51} & \textbf{74.21} & 62.43  & \textbf{65.87}  & 61.07  & \textbf{37.29}  & \textbf{60.17}   \\ 
    \midrule
        \multirow{6}{*}{\textbf{Llama2-13B}} & Dense & 0 & 4.89 & 80.52  & 72.14  & 79.36  & 77.36  & 49.23  & 71.72   \\ 
        ~ & SliceGPT & 21.52 & 23.95 & 74.32  & 65.59  & 60.71  & \textbf{68.52}  & \textbf{42.41}  & 62.31   \\ 
        ~ & LaCo & 24.37 & 13.97 & 72.42  & 59.27  & 60.44  & 54.34  & 34.56  & 56.21   \\ 
        ~ & RM & 24.37 & 10.08 & 73.72  & 66.61  & 66.80  & 66.12  & 41.98  & 63.05   \\ 
        ~ & ShortGPT & 24.37 & 20.06 & 72.74  & \textbf{70.80}  & 67.80  & 60.35  & 41.30  & 62.60   \\ 
        \rowcolor[gray]{.93} \cellcolor{white} ~ & \cellcolor{white}BlockPruner & \cellcolor{white}25.12 &  \textbf{8.16} & \textbf{76.93}  & 66.30  & \textbf{72.20}  & 65.82  & 41.38  & \textbf{64.53}   \\ 
    \midrule
        \multirow{5}{*}{\textbf{Baichuan2-7B}} & Dense & 0 & 6.04 & 77.48  & 68.27  & 72.18  & 72.98  & 42.75  & 66.73   \\ 
        ~ & LaCo & 21.57 & 26.46 & 68.28  & 58.56  & 51.50  & 52.90  & 28.50  & 51.95   \\ 
        ~ & RM & 21.57 & 189.78 & 59.96  & 52.33  & 30.87  & 38.17  & 23.63  & 40.99   \\ 
        ~ & ShortGPT & 21.57 & 31.05 & 63.71  & \textbf{62.67}  & 50.01  & 47.31  & 30.72  & 50.88   \\ 
        \rowcolor[gray]{.93} \cellcolor{white} ~ & \cellcolor{white}BlockPruner & \cellcolor{white}22.45 &  \textbf{15.38} & \textbf{69.75}  & 61.48  & \textbf{58.09}  & \textbf{58.08}  & \textbf{33.02}  & \textbf{56.08}   \\ 
    \midrule
        \multirow{5}{*}{\textbf{Baichuan2-13B}} & Dense & 0 & 6.66 & 78.84  & 70.40  & 75.23  & 74.07  & 47.70  & 69.25   \\ 
        ~ & LaCo & 22.68 & 27.07 & 70.89  & 58.01  & 54.00  & 57.11  & 32.94  & 54.59   \\ 
        ~ & RM & 22.68 & 17.70 & 68.99  & 67.88  & 63.78  & 57.49  & 37.54  & 59.14   \\ 
        ~ & ShortGPT & 22.68 & 20.69 & 69.31  & \textbf{68.27}  & 61.71  & 56.52  & 36.69  & 58.50   \\ 
        \rowcolor[gray]{.93} \cellcolor{white} ~ & \cellcolor{white}BlockPruner & \cellcolor{white}24.19 &  \textbf{15.36} & \textbf{71.44}  & 64.01  & \textbf{64.20}  & \textbf{59.81}  & \textbf{37.88}  & \textbf{59.47}   \\ 
    \midrule
        \multirow{5}{*}{\textbf{Qwen1.5-7B}} & Dense & 0 & 7.95 & 79.22  & 66.46  & 76.92  & 62.16  & 42.66  & 65.48   \\ 
        ~ & LaCo & 20.97 & 39.23 & 70.40  & 58.64  & 56.35  & 46.89  & 32.85  & 53.03   \\ 
        ~ & RM & 20.97 & 2026.31 & 67.36  & 49.88  & 42.00  & \textbf{54.17}  & 28.58  & 48.40   \\ 
        ~ & ShortGPT & 20.97 & 49.88 & 69.53  & \textbf{62.12}  & 58.87  & 43.60  & 32.17  & 53.26   \\ 
       \rowcolor[gray]{.93} \cellcolor{white} ~ & \cellcolor{white}BlockPruner & \cellcolor{white}21.83 &  \textbf{20.58} & \textbf{71.71}  & 55.56  & \textbf{59.31}  & 53.70  & \textbf{33.28}  & \textbf{54.71}   \\ 
    \midrule
        \multirow{5}{*}{\textbf{Qwen1.5-14B}} & Dense & 0 & 7.44 & 79.87  & 70.56  & 79.41  & 68.48  & 47.01  & 69.07   \\ 
        ~ & LaCo & 22.25 & 16.32 & 71.55  & 58.33  & 60.16  & 53.70  & 34.04  & 55.56   \\ 
        ~ & RM & 22.25 & 55.99 & 67.08  & 53.28  & 42.08  & 50.72  & 29.01  & 48.43   \\ 
        ~ & ShortGPT & 22.25 & 1237.21 & 58.60  & 55.96  & 36.16  & 38.09  & 34.81  & 44.72   \\ 
        \rowcolor[gray]{.93} \cellcolor{white} ~ & \cellcolor{white}BlockPruner & \cellcolor{white}23.72 &  \textbf{15.67} & \textbf{75.24}  & \textbf{61.48}  & \textbf{66.92}  & \textbf{59.51}  & \textbf{39.08}  & \textbf{60.45}   \\ 
    \bottomrule
    \end{tabular}
}
\caption{Zero-shot downstream task performance of various models using different pruning methods. ``Dense'' represents the original, unpruned models. ``PPL'' means the perplexity on Wikitext2. 
}
\label{tab:main_res}
\vspace{-0.3cm}
\end{table*}

We compare our method with several state-of-the-art structured pruning methods. 
The specific baseline methods include \textbf{SliceGPT} \citep{ashkboos2024slicegpt}, \textbf{LaCo} \citep{yang2024laco}, \textbf{ShortGPT} \citep{men2024shortgpt}, and \textbf{Relative Magnitude} \citep{samragh2023weight, men2024shortgpt}. SliceGPT achieves pruning by removing rows or columns corresponding to smaller principal components in the weight matrix. 
LaCo merges model layers from deep to shallow, using model output representations to calculate thresholds to avoid over-merging. ShortGPT eliminates redundant layers by calculating Block Influence.
Relative Magnitude (RM) uses \( ||\frac{f(x)}{x+f(x)}|| \) as an importance metric for layers, where \( f(.) \) represents the non-residual part of the Transformer layer, and employs the same pruning method as ShortGPT. 
For SliceGPT, we used the official implementation\footnote{As SliceGPT's official code does not support Baichuan2 and Qwen1.5, we only employ it on the Llama2 series models.}. For LaCo, we implemented it based on their code and controlled the number of pruned layers by adjusting the merging threshold. For ShortGPT and RM, we reproduced the results based on their manuscripts. More detailed implementation information is provided in Appendix \ref{sec:impl_detail}.

\paragraph{Data and GPUs.}
In our main experiment, we utilize the Alpaca dataset \citep{alpaca} to calculate importance scores. For our method, we employ only 256 samples to compute perplexity, and we discuss the influence of varying sample sizes in Section \ref{subsec:impact_data}. To ensure consistency, we use the same number of samples for ShortGPT and Relative Magnitude methods as shown in Appendix \ref{sec:impl_detail}. Moreover, the effect of sample size on ShortGPT and Relative Magnitude is detailed in Appendix \ref{sec:sens_data_size}.
All experiments are conducted on two RTX 4090 GPUs, and the execution times for different methods are reported in Appendix \ref{sec:time_cost}.

\paragraph{Evaluations.}

Following SliceGPT, we use LM Evaluation Harness \citep{eval-harness} for evaluation and validation on five well-known benchmarks: PIQA \citep{bisk2020piqa}, WinoGrande \citep{sakaguchi2021winogrande}, HellaSwag \citep{zellers2019hellaswag}, ARC-e and ARC-c \citep{clark2018think}. We also utilize Wikitext2 dataset \citep{merity2016pointer} for evaluating the perplexity after pruning. 
More comprehensive details of can be found in Appendix \ref{sec:eval_detail}.

\subsection{Main Results}

Prior studies \citep{yang2024laco, ashkboos2024slicegpt} have generally constrained the pruning ratio to approximately 25\%. In line with these studies, we also restricted the pruning ratio to this range in our main experiments.
Since it is challenging to achieve identical pruning ratios across different methods and models, we select the closest available pruning ratios for comparison. 

As shown in Table \ref{tab:main_res}, our BlockPruner method significantly outperforms previous structured pruning baselines in terms of average performance and achieves the best results across most benchmarks, even though the pruning ratios in our method are slightly higher than that of baselines. We also observe that Llama2-13B maintains better performance at higher pruning ratios compared to Llama2-7B, with Baichuan2 and Qwen1.5 exhibiting similar behavior. This suggests that as the model scale grows, so does the number of redundant blocks, allowing for more pruning space.

Furthermore, it's noteworthy that models with lower perplexity on the Wikitext2 dataset tend to perform better, which highlights the correlation between perplexity and model effectiveness. This further supports the validity of perplexity as a reliable metric for evaluating model performance.
Remarkably, although our method performs pruning searches on the Alpaca dataset, it achieves lower perplexity on the Wikitext2 dataset.

Finally, we observe that while approaches such as ShortGPT and Relative Magnitude result in a significant decline in model performance across different tasks, BlockPruner stands out by avoiding such drastic reductions. This suggests that our proposed block pruning method effectively mitigates performance degradation during the pruning process. 
Due to space constraints, we have moved the details of pruning baselines and comparisons across various pruning ratios to Appendix \ref{sec:vary_pr}. 

\section{Analyses}
\label{sec:analyses}
\subsection{Ablation Study}

To assess the influence of various key operations within the proposed pruning algorithm on its performance, we undertake a thorough ablation study across six models. In particular, we first remove all blocks with the lowest importance scores at once, without the iterative search procedure. 
Then, we substitute the fine-grained block pruning with a coarser-grained layer pruning approach. The results of these experiments are shown in Table \ref{tab:abla_res}.

The experimental findings highlight that solely relying on the perplexity metric without incorporating a search component can result in subpar pruning results and even performance deterioration. This phenomenon may stem from the intrinsic nature of perplexity, which, unlike other importance metrics focusing solely on local block influence, is inherently influenced by the interaction among multiple blocks due to its derivation from the model's output calculation. While perplexity aids in identifying redundant blocks within the model, it doesn't directly yield an optimal pruning sequence.

Furthermore, pruning at the layer level rather than the block level yields less robust performance. This observation indicates that the model contains fine-grained redundancies, and segmenting layers into smaller blocks for pruning allows for more efficient removal of this redundancy, thereby better preserving the model's capabilities. Additionally, we provide ablation experiments at higher sparsity levels, with results presented in Appendix \ref{sec:high_abla}.

\begin{table}[!t]
    \centering
\resizebox{0.9\linewidth}{!}{
    \begin{tabular}{l|l|c|c}
    \toprule
        \textbf{Model} & \textbf{Method} & \textbf{Ratio (\%)} & \textbf{Avg. Score } \\ 
    \midrule
        \multirow{3}{*}{\textbf{Llama2-7B}} & \textbf{BlockPruner} & 21.99 & \textbf{60.17}  \\ 
        ~ & - search & 20.95 & 55.89 (\textcolor{darkred}{-7.11\%})  \\ 
        ~ & - block & 21.02 & 58.63  (\textcolor{darkred}{-2.56\%})  \\
    \midrule
        \multirow{3}{*}{\textbf{Llama2-13B}} & \textbf{BlockPruner} & 25.12 & \textbf{64.53}  \\
        ~ & - search & 25.08 & 58.58  (\textcolor{darkred}{-9.21\%}) \\
        ~ & - block & 24.37 & 62.91  (\textcolor{darkred}{-2.51\%}) \\
    \midrule
        \multirow{3}{*}{\textbf{Baichuan2-7B}} & \textbf{BlockPruner} & 22.45 & \textbf{56.08}  \\
        ~ & - search & 22.39 & 38.81 (\textcolor{darkred}{-30.80\%})  \\
        ~ & - block & 21.57 & 54.76 (\textcolor{darkred}{-2.36\%})  \\
    \midrule
        \multirow{3}{*}{\textbf{Baichuan2-13B}} & \textbf{BlockPruner} & 24.19 & \textbf{59.47}  \\
        ~ & - search & 24.19 & 55.95 (\textcolor{darkred}{-5.92\%})  \\
        ~ & - block & 24.95 & 58.22 (\textcolor{darkred}{-2.10\%})  \\
    \midrule
        \multirow{3}{*}{\textbf{Qwen1.5-7B}} & \textbf{BlockPruner} & 21.83 & \textbf{54.71}  \\
        ~ & - search & 20.90 & 37.72 (\textcolor{darkred}{-31.06\%})  \\
        ~ & - block & 20.97 & 52.66 (\textcolor{darkred}{-3.75\%})  \\
    \midrule
        \multirow{3}{*}{\textbf{Qwen1.5-14B}} & \textbf{BlockPruner} & 23.72 & \textbf{60.45}  \\
        ~ & - search & 22.98 & 40.80 (\textcolor{darkred}{-32.51\%})  \\
        ~ & - block & 22.25 & 60.10 (\textcolor{darkred}{-0.58\%})  \\
    \bottomrule
    \end{tabular}
}
\vspace{-0.2cm}
\caption{Average score of ablation study of BlockPruner on downstream tasks. ``- search'' indicates dropping the iterative search procedure and directly removing blocks with the lowest importance score. ``- block'' means we substitute the fine-grained block pruning with
a coarser-grained layer pruning approach.}
\label{tab:abla_res}
\vspace{-0.3cm}
\end{table}

\subsection{Redundancies Between MHA and MLP}
\label{subsec:diff_block}
\begin{figure}[t!]
    \centering
    \includegraphics[width=0.98\linewidth]{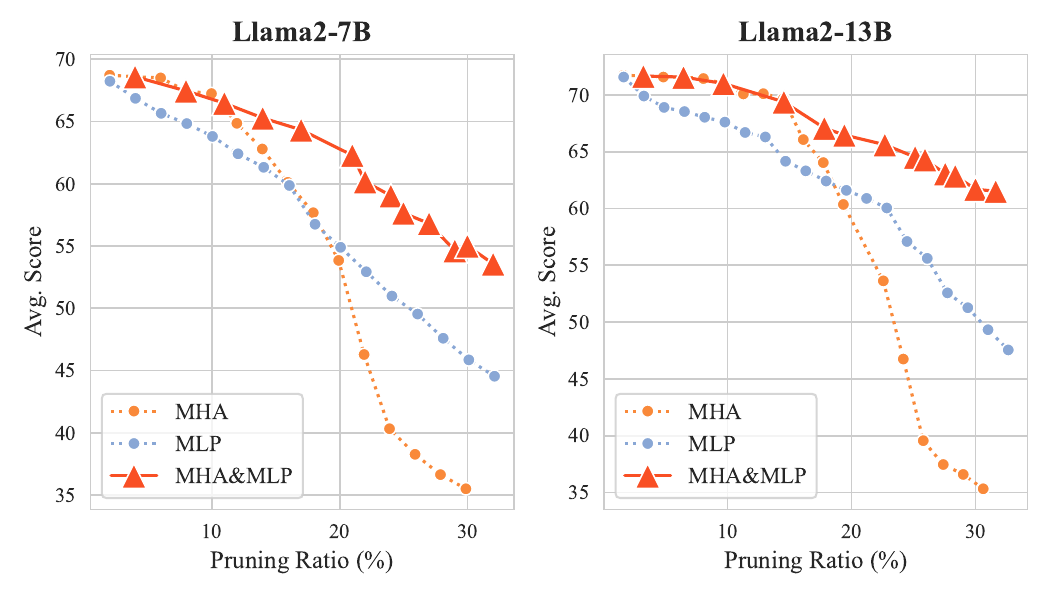}
    \caption{The impact of pruning MHA and MLP individually with different pruning
ratios on model performance. ``MHA\&MLP'' represents the original BlockPruner algorithm. Results show that MHA modules in LLMs are more redundant than MLP modules.}
    \label{fig:diff_block}
\vspace{-0.5cm}
\end{figure}

To investigate the significance and roles of the MHA and MLP modules in modern LLMs, we conduct pruning experiments focusing exclusively on MHA or MLP blocks. We apply this pruning strategy to two models of varying sizes, Llama2-7B and Llama2-13B, while keeping the pruning ratios below 33\%. The results illustrated in Figure \ref{fig:diff_block}  reveal several notable observations.

Before reaching a pruning ratio of 17\%, pruning only the MHA blocks results in less performance loss compared to pruning MLP blocks and even matches the performance of mixed pruning. This indicates that MHA modules in LLMs may possess greater redundancy than initially anticipated, whereas MLP modules are relatively less redundant. However, when the pruning ratio surpasses 17\%, further pruning of MHA blocks leads to a sharp decline in performance. This trend suggests that as pruning advances, the redundant MHA blocks are progressively removed, leaving only the crucial MHA blocks. Moreover, in the larger model, the sharp decline in performance occurs at higher pruning ratios, which is consistent with the finding that larger models contain more redundant blocks. Such redundancy may stem from factors like insufficient training, resulting in higher initial redundancy.

\begin{figure}
    \centering
    \includegraphics[width=0.95\linewidth]{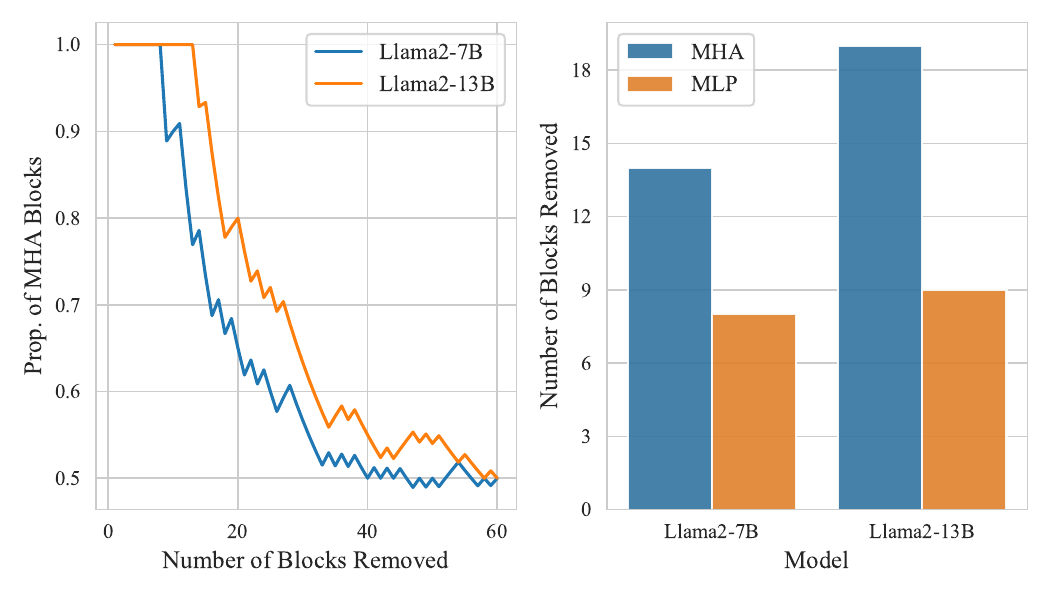}
    \caption{\textbf{Left}: The proportion of MHA blocks removed during the pruning process, relative to the total number of removed blocks. \textbf{Right}: The number of different blocks removed from models at a pruning ratio of 30\%.}
    \label{fig:attn_prop}
    \vspace{-0.3cm}
\end{figure}

We also examine the proportion of MHA blocks removed during pruning. Specifically, we present the number of MHA and MLP blocks removed at different pruning stages. In Figure \ref{fig:attn_prop} (left), we set the number of removed blocks to 60. In Figure \ref{fig:attn_prop} (right),  the models have 22 and 28 blocks removed, respectively, maintaining a pruning ratio of 30\%.

The results in Figure \ref{fig:attn_prop} (left) for both models reveal a consistent tendency to initially remove only MHA blocks. As the pruning process progresses and more blocks are removed, the proportion of MHA blocks being pruned follows a zigzag downward trend. Notably, the curve for Llama2-13B shifts to the right compared to Llama2-7B, suggesting that the larger model contains more redundant MHA blocks. This is further emphasized in Figure \ref{fig:attn_prop} (right), where, at the same pruning ratio, Llama2-13B prunes more MHA blocks than Llama2-7B. 
Additionally, given that our pruning method tends to remove more MHA blocks at equivalent pruning ratios, it can significantly reduce the usage of the key-value (KV) cache \citep{pope2023efficiently} in MHA, which potentially accelerate the inference process.
To validate this, we also conducted a comparison of the inference speed among various models obtained through different pruning methods, with the results detailed in Appendix \ref{sec:infer_speed}.

\subsection{Perplexity for Block Redundancy}
\label{subsec:diff_metrics}
In this section, we explore the impact of different block importance metrics. 
Generally, Block Influence (BI) and Relative Magnitude (RM) measure the importance of a block based solely on its input and output hidden states, thereby reflecting the block's local influence. In contrast, perplexity is derived from the model's output representations and thus can better measure a block's overall influence.

However, as indicated in the ablation study, using perplexity without the iterative search procedure leads to a significant decline in performance. 
This suggests that while perplexity alone may not be a strong block importance metric, our iterative search method allows for a more effective use of it.

\begin{figure}
    \centering
    \includegraphics[width=0.95\linewidth]{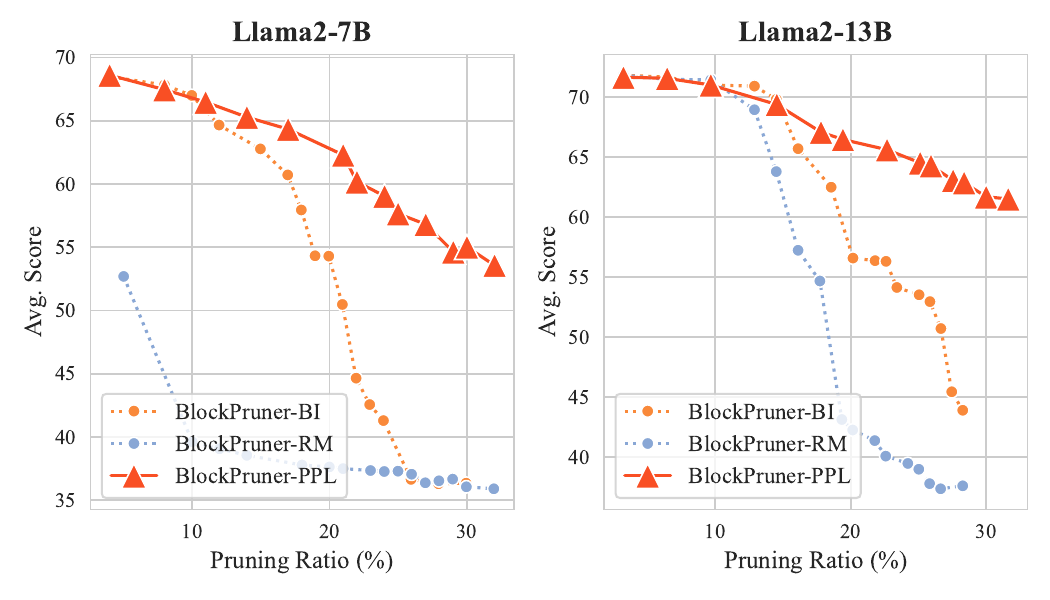}
        \vspace{-0.2cm}
    \caption{The impact of different block importance metrics on the pruning performance of BlockPruner}
    \label{fig:diff_metrics}
    \vspace{-0.2cm}
\end{figure}

As illustrated in Figure \ref{fig:diff_metrics}, when BI and RM are applied in dynamic pruning algorithms, they sometimes achieve performance comparable to perplexity at lower pruning ratios. However, as the pruning ratio increases, their limitations become evident, resulting in a sharp decline in model performance. This suggests that these local metrics do not adequately capture the impact of different blocks on the model's overall performance.

In summary, perplexity leverages global information to effectively measure block redundancy, especially when used with a dynamic pruning strategy. This combination captures the complex interactions among blocks. 
In contrast, local metrics like BI and RM are useful in specific scenarios but don't reflect the overall contribution of blocks to the model, particularly at higher pruning ratios.

\subsection{Impact of Data on Pruning}
\label{subsec:impact_data}
\begin{figure}
    \centering
    \includegraphics[width=0.95\linewidth]{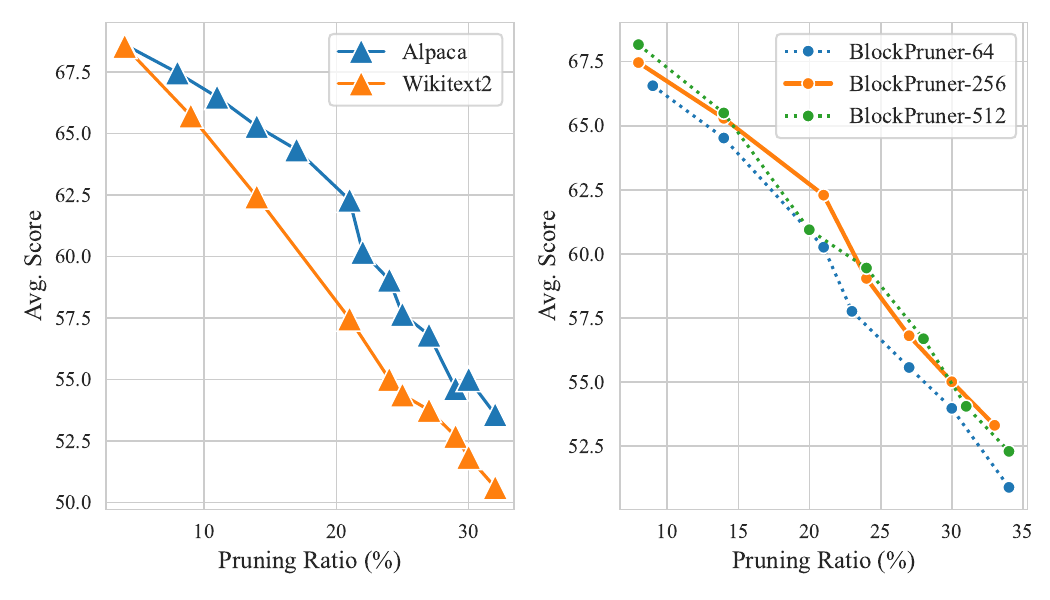}
    \caption{\textbf{Left}: The performance of BlockPruner on the Alpaca and Wikitext2 datasets using a calibration dataset of 256 samples. \textbf{Right}: Impact of sample sizes on BlockPruner's performance on Alpaca, with the numbers indicating the sample sizes used.}
    \label{fig:impact_data}
    \vspace{-0.5cm}
\end{figure}

In the work on SliceGPT \citep{ashkboos2024slicegpt}, the authors also used the Wikitext2 \citep{merity2016pointer} and Alpaca \citep{alpaca} datasets for pruning experiments. They observed that the Alpaca dataset often yielded better pruning results. In our study, we obtain similar findings. As shown in Figure \ref{fig:impact_data} (left), when pruning Llama2-7B, the performance across different pruning ratios is significantly higher when using the Alpaca dataset compared to Wikitext2. We hypothesize that this may be due to the Alpaca dataset being an instruction-following dataset, which is more closely aligned with downstream tasks. This suggests that the choice of dataset has a significant impact on the final pruning performance of the model. 

To determine the appropriate sample size and analyze its impact on the pruning performance of BlockPruner, we extract varying numbers of instances from the Alpaca dataset and conduct pruning experiments using Llama2-7B. The results presented in Figure \ref{fig:impact_data} (right) indicate that increasing the sample size beyond 256 yields no significant improvement in the pruning effect of BlockPruner. Therefore, we set the number of samples to 256.

\section{Conclusion}
In this work, we introduce BlockPruner, a novel structured pruning approach for efficiently pruning LLMs. BlockPruner decomposes Transformer layers into two minimal residual blocks and leverages a block importance metric based on perplexity in conjunction with an iterative pruning search algorithm, where the two components work together to progressively eliminate redundant blocks.
Extensive experiments across various models show that our method outperforms other baselines in post-pruning performance. Our findings uncover fine-grained block redundancy in LLMs, highlighting significant differences in redundancy levels across different block types. We hope our work contributes to a deeper understanding of the importance of different blocks within LLMs.

\section*{Limitations}
Our current work has three potential limitations. First, while perplexity serves as a useful indicator of block importance, it may not be the optimal metric. Second, while our proposed pruning search algorithm is effective, other combinatorial optimization algorithms might identify superior pruning sequences. Lastly, due to constraints in computational resources, we did not apply our method to prune larger models. Nevertheless, our approach is highly scalable and readily adaptable for pruning larger models in future research.

\section*{Ethics Statement}
The aim of this study is to provide a generalizable pruning method for large language models. All models and datasets used in our experiments are publicly accessible and do not contain any private information. We strictly adhere to the usage policies of these resources and utilize them solely for research purposes.

\section*{Acknowledgments}
This work was supported by the National Natural Science Foundation of China (No. 62176270) and the Guangdong Basic and Applied Basic Research Foundation (No. 2023A1515012832).

\bibliography{acl_latex}

\appendix
\newpage
\section{Details of Implementations}
\label{sec:impl_detail}
In this section, we detail our experimental setup.  We sampled from the Alpaca dataset with a fixed random seed of 42. For SliceGPT, we followed the original paper's configuration, using 1024 samples, a sparsity ratio set at 30\%, and a maximum sequence length of 2048. For ShortGPT, RM, and BlockPruner, we sampled 256 samples from the dataset, with the same maximum sequence length of 2048. For LaCo, we adjusted the merging threshold using the provided code and data to achieve the corresponding pruning ratio.

\section{Details of Datasets}

\subsection{Pruning Datasets}

\textbf{Alpaca} \citep{alpaca} is a general instruction-following dataset containing 52,000 questions. Each sample comprises three fields: instruction, input, and response. We selected 10\% of the dataset and utilized 256 samples for the main experiments. Perplexity calculation was performed uniformly across all text in the samples without differentiation between fields.

\subsection{Evaluation Datasets}

All downstream task datasets were partitioned and evaluated using the default configuration of LM Evaluation Harness.

\textbf{Wikitext-2} \citep{merity2016pointer} is a collection of over 100 million tokens extracted from verified Good and Featured articles on Wikipedia. This dataset is commonly used to measure the quality of a model's text generation.  We employed samples from the pre-split test set for calculating perplexity.

\textbf{PIQA} \citep{bisk2020piqa} is a dataset designed to evaluate natural language models' understanding of physical commonsense. It employs a multiple-choice format where the model selects the most appropriate solution from two options given a goal.

\textbf{WinoGrande} \citep{sakaguchi2021winogrande} is an extensive dataset to evaluate models' commonsense reasoning capabilities. 
It comprises 44,000 questions. The dataset features fill-in-the-blank tasks with binary options, aiming to select the correct option for a given sentence that requires commonsense reasoning.

\textbf{HellaSwag} \citep{zellers2019hellaswag} is also a dataset designed to assess models' commonsense reasoning abilities, specifically to highlight the limitations of current models in handling commonsense natural language reasoning tasks. Despite being trivial for humans (with >95\% accuracy), the dataset presents significant difficulties for models. The evaluation is conducted using four-way multiple-choice questions.

\textbf{ARC} \citep{clark2018think} dataset comprises 7,787 multiple-choice science exam questions sourced from various origins. Each question typically offers four answer options. These questions are categorized into two distinct difficulty sets: 2,590 questions for Challenge Set and 5,197 for Easy Set.

\section{Details of Evaluations}
\label{sec:eval_detail}
Ensuring a fair and comprehensive comparison, we employed the same version of the LM Evaluation Harness as used in the SliceGPT experiments, obtaining evaluation scores under identical experimental configurations. These scores closely match those reported in the SliceGPT paper, as detailed in Table \ref{tab:slice_comp}. For consistency, we present our reproduced results in the main experiments.

\begin{table}[t]
    \centering
\resizebox{0.9\linewidth}{!}{
    \begin{tabular}{llcc}
    \toprule
        \textbf{Model} & \textbf{Method}  & \textbf{Ratio(\%)} & \textbf{Avg.Score}  \\ 
    \midrule
        \multirow{2}{*}{\textbf{Llama2-7B}} & $\mathrm{SliceGPT}$ & 21.45 & 57.93  \\ 
        ~ & $\mathrm{SliceGPT}^*$ & 21.45 & 57.83  \\
    \midrule
        \multirow{2}{*}{\textbf{Llama2-13B}} & $\mathrm{SliceGPT}$ & 21.52 & 62.34  \\ 
        ~ & $\mathrm{SliceGPT}^*$ & 21.52 & 62.31  \\ 
    \bottomrule
    \end{tabular}
}
\caption{Comparison of average performance on downstream tasks between the official SliceGPT results and our reproduced results (indicated by ``$*$'' for our results).}
\label{tab:slice_comp}
\end{table}

For evaluating the performance of pruned models on downstream tasks, we utilized five multiple-choice QA datasets: PIQA, WinoGrande, HellaSwag, ARC-e, and ARC-c. Additionally, to assess text generation quality, we calculated perplexity using the test set of the Wikitext2 dataset. 
For the downstream task evaluations, we adhered to the default evaluation parameters and zero-shot settings, with a batch size set to 1. For perplexity calculations, the maximum text length was set to 2048, maintaining a batch size of 1 as well.

\section{Perplexity and JS Divergence in Block Evaluation}
\label{sec:ppl_vs_js}
Recent work such as FINERCUT~\citep{zhang2024finercut} proposes a fine-grained pruning algorithm that evaluates block importance using the \textbf{JS divergence} between the output distributions of the original and pruned models. While this metric captures distributional shifts, it overlooks the semantic fluency and coherence of generated text—key aspects for maintaining the practical utility of LLMs. In contrast, we adopt \textbf{perplexity (PPL)} as a global importance metric derived from sequence-level log-likelihood, which more directly reflects how pruning impacts output quality and fluency in models.

To validate this perspective, we conducted experiments using both metrics across various model scales and pruning ratios. The results, summarized in Table \ref{tab:ppl_vs_js}, indicate that PPL consistently outperforms JS divergence under different configurations. These findings demonstrate that PPL better reflects the fluency and quality of the pruned model’s outputs, reinforcing its suitability as a block importance metric for LLM pruning.
\begin{table}[h]
    \centering
\resizebox{0.99\linewidth}{!}{
    \begin{tabular}{llcc}
    \toprule
        \textbf{Model} & \textbf{Metric}  & \textbf{Ratio(\%)} & \textbf{Avg.Score}  \\ 
    \midrule
        \multirow{2}{*}{\textbf{Llama2-7B}} & PPL & 16.98/24.99/30.00/37.02 & \textbf{64.33}/\textbf{57.65}/\textbf{55.01}/\textbf{49.36} \\ 
        ~ & JS & 14.99/26.00/29.01/37.02 & 63.76/57.08/53.71/49.04 \\
    \midrule
        \multirow{2}{*}{\textbf{Llama2-13B}} & PPL & 14.54/25.12/31.62/34.88 & \textbf{69.38}/\textbf{64.52}/\textbf{61.52}/\textbf{60.13} \\ 
        ~ & JS & 14.54/26.75/32.45/34.88 & 69.28/64.36/61.00/59.61 \\
    \bottomrule
    \end{tabular}
}
\caption{Comparison of PPL and JS divergence across different pruning ratios and model scales.}
\label{tab:ppl_vs_js}
\end{table}

\section{Ablation at Higher Sparsity}
\label{sec:high_abla}
BlockPruner is motivated by the goal of preserving model performance more effectively through fine-grained block pruning. Evaluating how block pruning performs at different levels of granularity, particularly under higher sparsity, is crucial for supporting our motivations and claims.
In light of this, we conducted ablation experiments with higher sparsity ratios on Llama2-7B and Llama2-13B models.
The results, shown in Table \ref{tab:high_spar}, confirm that our approach remains effective, further validating the motivations behind BlockPruner.
\begin{table}[H]
    \centering
\resizebox{0.99\linewidth}{!}{
    \begin{tabular}{llccc}
    \toprule
        \textbf{Model} & \textbf{Unit}  & \textbf{Ratio(\%)} & \textbf{PPL($\downarrow$)} & \textbf{Avg.Score}  \\ 
    \midrule
        \multirow{2}{*}{\textbf{Llama2-7B}} & Block & 30.00/37.02/43.03 & \textbf{16.28}/27.47/\textbf{49.65} & \textbf{55.02}/\textbf{49.36}/\textbf{46.95} \\ 
        ~ & Layer & 30.03/36.04/42.05 & 16.58/\textbf{27.05}/60.85 & 53.04/47.32/43.91 \\
    \midrule
        \multirow{2}{*}{\textbf{Llama2-13B}} & Block & 31.62/36.52/41.39 & \textbf{9.64}/\textbf{12.54}/\textbf{17.02} & \textbf{61.52}/\textbf{59.34}/\textbf{54.15} \\ 
        ~ & Layer & 31.68/36.56/41.43 & 10.48/13.29/18.04 & 59.31/56.17/51.61 \\
    \bottomrule
    \end{tabular}
}
\caption{Average score of BlockPruner at different pruning granularities under higher sparsity.}
\label{tab:high_spar}
\end{table}

\section{Inference Speed after Pruning}
\label{sec:infer_speed}
In this section, we evaluate the inference speed by measuring the time required to generate 128 tokens using models obtained from different pruning methods, all employing KV caches for efficient decoding.
Each configuration is repeated 20 times to ensure statistically robust results, and we report the average inference time across runs.

As shown in Table~\ref{tab:infer_speed}, our method consistently achieves the greatest speedup at comparable pruning ratios. 
This improvement stems from the fact that our approach prunes a greater proportion of MHA blocks at the same overall pruning ratio compared to other methods, leading to a substantial reduction in KV cache usage, which directly accelerates the inference process of LLMs.
\begin{table}[htbp]
    \centering
\resizebox{0.99\linewidth}{!}{
    \begin{tabular}{llccc}
    \toprule
        \textbf{Model} & \textbf{Method}  & \textbf{Ratio(\%)} & \textbf{Inference Time (ms)} & \textbf{Speedup}  \\ 
    \midrule
        \multirow{4}{*}{\textbf{Llama2-7B}} & Original & 0.00 & 4044.30 & 1.00 \\ 
        ~ & BlockPruner & 24.00 & \textbf{2747.88} & \textbf{1.47} \\ 
        ~ & SliceGPT & 24.47 & 3226.68 & 1.25 \\ 
        ~ & ShortGPT & 24.03 & 3094.36 & 1.31 \\ 
    \midrule
        \multirow{4}{*}{\textbf{Llama2-13B}} & Original & 0.00 & 7285.73 & 1.00 \\ 
        ~ & BlockPruner & 21.05 & \textbf{3873.20} & \textbf{1.88} \\ 
        ~ & SliceGPT & 21.52 & 4099.08 & 1.78 \\ 
        ~ & ShortGPT & 21.93 & 4111.65 & 1.77 \\ 
    \bottomrule
    \end{tabular}
}
\caption{The inference speed differences among models obtained using different pruning methods, where ``Original'' denotes the unpruned model.}
\label{tab:infer_speed}
\end{table}

\section{Time Costs of Pruning Methods}
\label{sec:time_cost}
Our approach relies on PPL to determine block importance, which requires calculating PPL before pruning, making it challenging to design a more efficient pruning strategy. We have compared other block importance metrics (in Section \ref{subsec:diff_metrics}) but found that PPL still preserves the model's performance best. Moreover, since our method better maintains model performance and pruning is one-time without increasing subsequent inference overhead, so we believe the trade-off is worthwhile. The comparison results of pruning times between BlockPruner and other methods are presented in Table \ref{tab:time_cost}.

\begin{table}[H]
    \centering
\resizebox{0.99\linewidth}{!}{
    \begin{tabular}{lccccc}
    \toprule
        \textbf{Model} & \textbf{BlockPruner}  & \textbf{SliceGPT} & \textbf{ShortGPT} & \textbf{RM} & \textbf{LaCo}  \\ 
    \midrule
        \textbf{Llama2-7B} & 45 mins & 2 hours 9 mins & 2 mins & < 1 mins & 2 mins \\
    \midrule
        \textbf{Llama2-13B} & 2 hours 27 mins & 3 hours 30 mins & 2 mins & < 1 mins & 24 mins \\
    \bottomrule
    \end{tabular}
}
\caption{Execution time of BlockPruner and other pruning methods in the main experiment.}
\label{tab:time_cost}
\end{table}

\section{Post-training after Pruning}
\begin{figure}[t]
    \centering
    \includegraphics[width=0.95\linewidth]{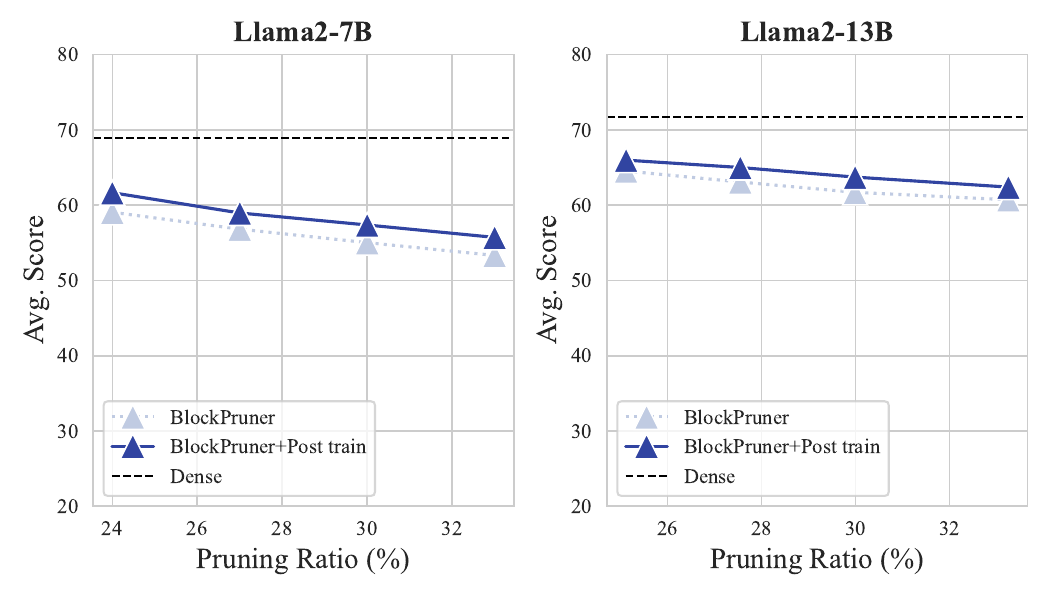}
    \caption{Average score of BlockPruner with varying pruning ratios before and after post-training.}
    \label{fig:post_train}
\end{figure}

We sampled 8,000 instances from the Alpaca dataset and conducted post-training on the pruned Llama2-7B and Llama2-13B models obtained via BlockPruner using LoRA. All linear layers, excluding the embedding layer and the language model head, were trained. The LoRA rank and LoRA $\alpha$ parameters were set to 32 and 10, respectively, with a learning rate of 2e-4 and a batch size of 1. Additionally, we configured the gradient accumulation steps to 4 and employed a linear learning rate scheduler.  
We controlled the pruning ratios within the range of 24\% to 33\%. The results are shown in Figure \ref{fig:post_train}. It can be seen that after training, our models showed further improvement at different pruning ratios. The Llama2-7B and Llama2-13B models recovered to 89\% and 92\% of the performance of the unpruned models, respectively, when pruned by approximately 1/4.

\section{Sensitivity to Sample Size}
\label{sec:sens_data_size}
\begin{figure*}
    \centering
    \includegraphics[width=0.99\linewidth]{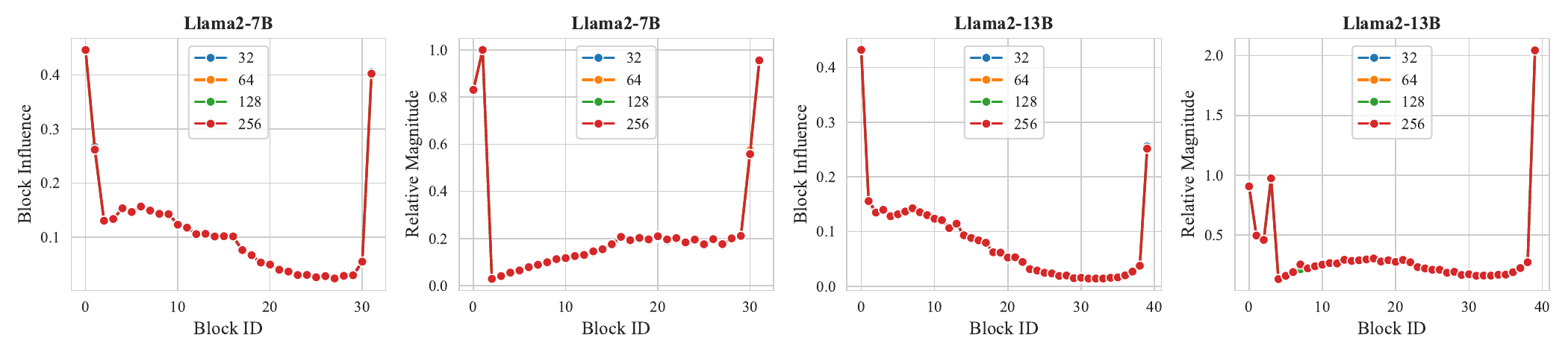}
    \caption{The changes in Block Influence and Relative Magnitude of the model under different sample sizes.}
    \label{fig:bi_rm_size}
\end{figure*}

ShortGPT uses Block Influence as the importance metric for layers, while RM uses Relative Magnitude. The former calculates the similarity between the input and output hidden states of a layer, while the latter utilizes the input and the non-residual part of the output. In our preliminary experiments, we found that these two metrics are not sensitive to sample size. We sampled different numbers of instances from the test set of the Alpaca dataset to observe their impact on these metrics, and the results are shown in Figure \ref{fig:bi_rm_size}. We can see that all the lines almost overlap, indicating that Block Influence and Relative Magnitude are not sensitive to the sample size. We speculate that this may be due to the limited information provided by the changes in the input and output of a single layer.

\begin{figure*}
    \centering
    \includegraphics[width=0.99\linewidth]{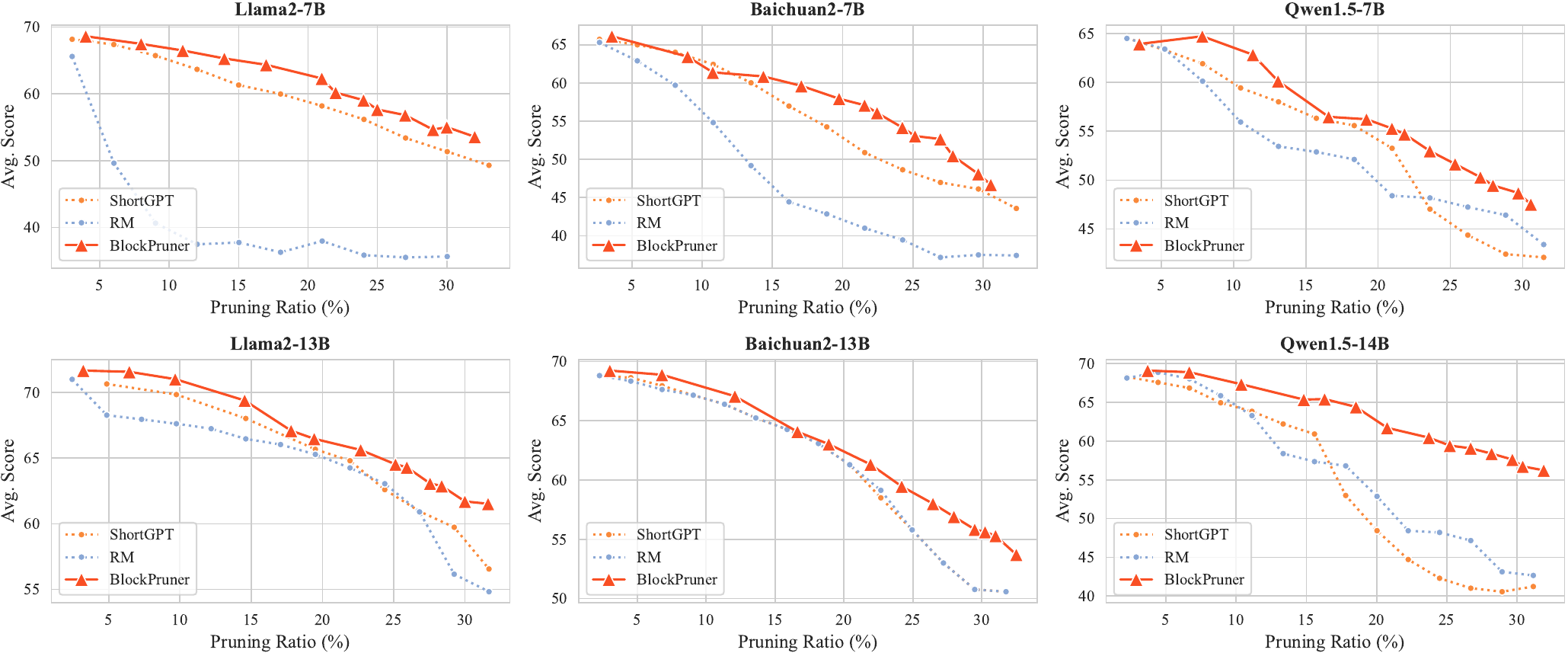}
    \caption{Average score of BlockPruner with varying pruning ratios compared with ShortGPT and RM.}
    \label{fig:diff_model_pr_lines}
\end{figure*}

\section{Retention of Reasoning Ability After Pruning}
To further assess the effectiveness of BlockPruner on reasoning-intensive scenarios, we evaluate its zero-shot performance on the AQuA-RAT dataset~\citep{ling-etal-2017-program}, a benchmark designed for algebraic word problem solving with rationales. We compare BlockPruner against several strong pruning baselines using the LLaMA2-7B and LLaMA2-13B models.

As summarized in Table~\ref{tab:aqua}, BlockPruner consistently maintains competitive performance, achieving accuracy close to the unpruned models and outperforming alternative pruning strategies such as ShortGPT, Relative Magnitude (RM), and LaCo. These results demonstrate that our method preserves the reasoning capabilities of LLMs even under significant structural compression.

\begin{table}[H]
\centering
\resizebox{0.8\linewidth}{!}{
\begin{tabular}{lcc}
\toprule
\textbf{Method} & \textbf{LLaMA2-7B} & \textbf{LLaMA2-13B} \\
\midrule
Original & 31.89 & 30.31 \\
BlockPruner & \textbf{29.92} & \textbf{29.53} \\
ShortGPT & 29.13 & 27.95 \\
RM & \textbf{29.92} & 26.77 \\
LaCo & 24.80 & 27.95 \\
\bottomrule
\end{tabular}
}
\caption{Zero-shot accuracy (\%) on AQuA-RAT for models pruned with different methods.
The pruning ratios for all methods are consistent with those used in the main experiments.}
\label{tab:aqua}
\end{table}

\section{Generalization to New Model Series}
To evaluate the generalization capability of BlockPruner to newly released model series, we conduct additional experiments on two recent architectures: Mistral-7B-v0.3 \citep{jiang2023mistral} and LLaMA3-8B \citep{grattafiori2024llama}. We compare our method against ShortGPT and Relative Magnitude (RM) across five pruning ratios, reporting average downstream task scores on each.

As shown in Table~\ref{tab:new_models}, BlockPruner consistently achieves superior performance on both model families across all sparsity levels. The performance gap becomes more pronounced as the pruning ratio increases, suggesting that our fine-grained pruning approach is particularly robust under high-compression regimes. These results confirm the effectiveness and adaptability of BlockPruner beyond the models evaluated in the main paper.

\begin{table}[H]
    \centering
\resizebox{0.99\linewidth}{!}{
    \begin{tabular}{llcc}
    \toprule
        \textbf{Model} & \textbf{Metric}  & \textbf{Ratio(\%)} & \textbf{Avg.Score}  \\ 
    \midrule
        \multirow{3}{*}{\textbf{Mistral-7B-v0.3}} & BlockPruner &
        25.12/27.55/29.98/36.00 &
        \textbf{58.02}/\textbf{57.02}/\textbf{56.10}/\textbf{49.18}\\
        ~ & ShortGPT & 24.07/27.08/30.09/36.11 & 57.40/53.26/42.79/41.41 \\
        ~ & RM & 24.07/27.08/30.09/36.11 & 37.75/35.89/36.49/35.52 \\
    \midrule
    \multirow{3}{*}{\textbf{ LLaMA3-8B}} & BlockPruner & 24.86/27.58/30.30/35.73 & \textbf{52.36}/\textbf{50.08}/\textbf{46.38}/\textbf{45.41} \\
    ~ & ShortGPT & 24.45/27.16/29.88/35.31 & 41.97/44.51/43.71/42.85 \\
    ~ & RM & 24.45/27.16/29.88/35.31 & 39.34/38.90/37.48/36.80 \\
    \bottomrule
    \end{tabular}
}
\caption{Downstream task performance on recently released model families. BlockPruner consistently maintains stronger performance across pruning ratios.}
\label{tab:new_models}
\end{table}

\section{Varying Pruning Ratios}
\label{sec:vary_pr}
To broadly demonstrate the superiority of our method, we present the pruning effects of BlockPruner, ShortGPT, and Relative Magnitude on six representative large models at different pruning ratios. As depicted in Figure \ref{fig:diff_model_pr_lines}, our method effectively minimizes performance loss throughout the pruning process, avoiding any sudden drops in performance. In contrast, RM exhibits significant instability and is prone to performance collapse. ShortGPT performs relatively well, but in the pruning experiments on Qwen1.5-14B, it also leads to severe performance degradation at higher pruning ratios. Overall, the advantages of our method become more pronounced as both model size and pruning ratio increase.

\newpage
\section{Evaluation on Additional Datasets}
We extended the primary experiment by incorporating four additional well-established evaluation datasets: SWAG \citep{zellers-etal-2018-swag}, TruthfulQA \citep{lin-etal-2022-truthfulqa}, OpenBookQA \citep{Mihaylov2018CanAS}, and RACE \citep{lai-etal-2017-race}. As illustrated in Table \ref{tab:expanded_res}, our proposed method consistently surpasses previous pruning baselines across this broader range of datasets, further demonstrating its effectiveness and generalization capability.
\begin{table*}[t]
    \centering
\resizebox{0.99\textwidth}{!}{
    \begin{tabular}{llccccccccccc}
    \toprule
        \textbf{Model} & \textbf{Method} & \textbf{Ratio (\%)} &\textbf{PIQA} & \textbf{WinoGrande} & \textbf{HellaSwag} & \textbf{ARC-e} & \textbf{ARC-c} & \textbf{TruthfulQA} & \textbf{RACE} & \textbf{SWAG} & \textbf{ObenBookQA} & \textbf{Avg. Score} \\ 
    \midrule
        \multirow{6}{*}{\textbf{Llama2-7B}} & Dense & 0 & 79.05  & 69.06  & 75.99  & 74.54  & 46.16 & 25.34 & 39.43 & 76.65 & 44.00  & 58.91   \\ 
          ~ & SliceGPT & 21.45 & 72.42  & 59.91  & 56.04  & \textbf{63.64}  & 37.12 & 25.34 & \textbf{37.22} & 61.57 & 33.20 & 49.61   \\ 
          ~ & LaCo & 21.02 & 68.34  & 60.46  & 54.08  & 55.39  & 35.84  & \textbf{28.40} & 29.57 & 61.75 & \textbf{39.80} & 48.18   \\ 
          ~ & RM & 21.02 & 54.46  & 49.25  & 29.22  & 34.43  & 22.53  & 26.07 & 22.58 & 38.60 & 27.60 & 33.86   \\ 
          ~ & ShortGPT & 21.02 & 70.24  & \textbf{65.90}  & 62.63  & 56.06  & 36.09 & 26.81 & 34.07 & 64.84 & 37.20 & 50.43  \\ 
          \rowcolor[gray]{.93} \cellcolor{white} ~ & \cellcolor{white}BlockPruner & \cellcolor{white}21.99 & \textbf{74.21} & 62.43  & \textbf{65.87}  & 61.07  & \textbf{37.29}  & 22.03 & 34.83 & \textbf{69.81} & 37.20 & \textbf{51.64}   \\ 
    \midrule
        \multirow{6}{*}{\textbf{Llama2-13B}} & Dense & 0 & 80.52  & 72.14  & 79.36  & 77.36  & 49.23  & 26.07 & 40.77 & 78.04 & 45.40 & 60.99   \\ 
        ~ & SliceGPT & 21.52 & 74.32  & 65.59  & 60.71  & \textbf{68.52}  & \textbf{42.41}  & 24.72 & 37.42 & 65.61 & 39.80 & 53.23   \\ 
        ~ & LaCo & 24.37 & 72.42  & 59.27  & 60.44  & 54.34  & 34.56  & 23.62 & 31.87 & 67.93 & \textbf{41.00} & 49.49   \\ 
        ~ & RM & 24.37 & 73.72  & 66.61  & 66.80  & 66.12  & 41.98  & 20.81 & 38.28 & 68.08 & 38.40 & 53.42  \\ 
        ~ & ShortGPT & 24.37 & 72.74  & \textbf{70.80}  & 67.80  & 60.35  & 41.30  & 24.60 & 37.80 & 68.67 & \textbf{41.00} & 53.90   \\ 
        \rowcolor[gray]{.93} \cellcolor{white} ~ & \cellcolor{white}BlockPruner & \cellcolor{white}25.12 & \textbf{76.93}  & 66.30  & \textbf{72.20}  & 65.82  & 41.38  & \textbf{24.97} & \textbf{38.85} & \textbf{72.94} & 40.60 & \textbf{55.55}  \\ 
    \midrule
        \multirow{5}{*}{\textbf{Baichuan2-7B}} & Dense & 0 & 77.48  & 68.27  & 72.18  & 72.98  & 42.75  & 23.01 & 38.76 & 75.26 & 40.00 & 56.74   \\ 
        ~ & LaCo & 21.57 & 68.28  & 58.56  & 51.50  & 52.90  & 28.50  & 21.42 & 31.10 & 62.37 & \textbf{33.60} & 45.36  \\ 
        ~ & RM & 21.57 & 59.96  & 52.33  & 30.87  & 38.17  & 23.63  & \textbf{25.09} & 22.01 & 47.38 & 27.40 & 36.32   \\ 
        ~ & ShortGPT & 21.57 & 63.71  & \textbf{62.67}  & 50.01  & 47.31  & 30.72  & 24.60 & 30.62 & 55.81 & 31.20 & 44.07   \\ 
        \rowcolor[gray]{.93} \cellcolor{white} ~ & \cellcolor{white}BlockPruner & \cellcolor{white}22.45 & \textbf{69.75}  & 61.48  & \textbf{58.09}  & \textbf{58.08}  & \textbf{33.02}  & 20.81 & \textbf{33.21} & \textbf{64.95} & 32.20 & \textbf{47.95}   \\ 
    \midrule
        \multirow{5}{*}{\textbf{Baichuan2-13B}} & Dense & 0 & 78.84  & 70.40  & 75.23  & 74.07  & 47.70  & 26.93 & 41.15 & 76.87 & 43.60 & 59.42   \\ 
        ~ & LaCo & 22.68 & 70.89  & 58.01  & 54.00  & 57.11  & 32.94  & 20.69 & 29.38 & \textbf{67.79} & 33.80 & 47.18   \\ 
        ~ & RM & 22.68 & 68.99  & 67.88  & 63.78  & 57.49  & 37.54  & 25.46 & \textbf{36.84} & 64.50 & 33.80 & 50.70   \\ 
        ~ & ShortGPT & 22.68 & 69.31  & \textbf{68.27}  & 61.71  & 56.52  & 36.69  & \textbf{26.93} & 36.27 & 64.14 & 34.40 & 50.47   \\ 
        \rowcolor[gray]{.93} \cellcolor{white} ~ & \cellcolor{white}BlockPruner & \cellcolor{white}24.19 & \textbf{71.44}  & 64.01  & \textbf{64.20}  & \textbf{59.81}  & \textbf{37.88}  & 23.50 & 36.75 & 67.43 & \textbf{35.40} & \textbf{51.16}   \\ 
    \midrule
        \multirow{5}{*}{\textbf{Qwen1.5-7B}} & Dense & 0 & 79.22  & 66.46  & 76.92  & 62.16  & 42.66  & 34.76 & 42.11 & 76.22 & 41.60 & 58.01   \\ 
        ~ & LaCo & 20.97 & 70.40  & 58.64  & 56.35  & 46.89  & 32.85  & 25.34 & 32.92 & \textbf{63.43} & \textbf{37.20} & 47.11  \\ 
        ~ & RM & 20.97 & 67.36  & 49.88  & 42.00  & \textbf{54.17}  & 28.58  & 21.66 & 23.54 & 58.32 & 35.40 & 42.32   \\ 
        ~ & ShortGPT & 20.97 & 69.53  & \textbf{62.12}  & 58.87  & 43.60  & 32.17  & \textbf{32.07} & 31.00 & 57.72 & 33.40 & 46.72   \\ 
       \rowcolor[gray]{.93} \cellcolor{white} ~ & \cellcolor{white}BlockPruner & \cellcolor{white}21.83 & \textbf{71.71}  & 55.56  & \textbf{59.31}  & 53.70  & \textbf{33.28}  & 25.70 & \textbf{34.74} & 61.32 & 33.80 & \textbf{47.68}   \\ 
    \midrule
        \multirow{5}{*}{\textbf{Qwen1.5-14B}} & Dense & 0 & 79.87  & 70.56  & 79.41  & 68.48  & 47.01  & 35.86 & 41.05 & 76.72 & 43.60 & 60.28   \\ 
        ~ & LaCo & 22.25 & 71.55  & 58.33  & 60.16  & 53.70  & 34.04  & 22.28 & \textbf{33.78} & 65.79 & 35.00 & 48.29   \\ 
        ~ & RM & 22.25 & 67.08  & 53.28  & 42.08  & 50.72  & 29.01  & 26.44 & 27.08 & 58.64 & 32.40 & 42.97   \\ 
        ~ & ShortGPT & 22.25 & 58.60  & 55.96  & 36.16  & 38.09  & 34.81  & 27.05 & 26.99 & 39.89 & 31.40 & 38.77   \\ 
        \rowcolor[gray]{.93} \cellcolor{white} ~ & \cellcolor{white}BlockPruner & \cellcolor{white}23.72 & \textbf{75.24}  & \textbf{61.48}  & \textbf{66.92}  & \textbf{59.51}  & \textbf{39.08}  & \textbf{30.60} & \textbf{33.78} & \textbf{67.39} & \textbf{38.20} & \textbf{52.47}   \\ 
    \bottomrule
    \end{tabular}
}
\caption{Zero-shot downstream task performance of various models using varied pruning methods. ``Dense'' denotes the original, unpruned models. All evaluations are conducted with the same configuration to ensure comparability.}
\label{tab:expanded_res}
\vspace{-0.3cm}
\end{table*}

\end{document}